\documentclass{article}

\usepackage[utf8]{inputenc}
\usepackage{graphicx}
\usepackage{textgreek}
\usepackage{float}
\usepackage{indentfirst}
\usepackage{multirow}
\usepackage{caption}
\usepackage{amsmath}
\usepackage{amssymb}

\title{AI-based Layout-to-Image Dataset Generation for Lithography Defect Detection}
\title{A Physics-Constrained, Design-Driven Methodology for Defect Dataset Generation in Optical Lithography}

\author{%
  \begin{minipage}{0.9\textwidth}
    \centering
    Yuehua Hu$^{1,2,\dagger}$,
    Jiyeong Kong$^{1,3,\dagger}$,
    Dong-yeol Shin$^{1}$,
    Jaekyun Kim$^{2,*}$,
    Kyung-Tae Kang$^{1,*}$\\[6pt]
    \small
    $^{1}$Autonomous Manufacturing \& Process R\&D Department, Korea Institute of Industrial Technology (KITECH),\\
    Sangnok-gu, Ansan-si, Gyeonggi-do 15588, Korea\\
    $^{2}$Department of Photonics and Nanoelectronics, Hanyang University, Ansan-si, Gyeonggi-do 15588, Korea\\
    $^{3}$Micro/Nano System Department, Korea University, 145 Anam-ro, Seongbuk-gu, Seoul 02841, Korea\\[4pt]
    $^{\dagger}$These authors contributed equally to this work.\\
    $^{*}$Co-corresponding authors: \texttt{jeakyunkim@hanyang.ac.kr}, \texttt{ktkang@kitech.re.kr}
  \end{minipage}
}

\date{December 2025}
\usepackage[a4paper, margin=2cm]{geometry}
\begin{document}
\maketitle

\begin{abstract}
The efficacy of Artificial Intelligence (AI) in micro/nano manufacturing is fundamentally constrained by the scarcity of high-quality and physically grounded training data for defect inspection. Lithography defect data from semiconductor industry are rarely accessible for research use, resulting in a severe shortage of publicly available datasets. To address this bottleneck in lithography, this study proposes a novel methodology for generating large-scale, physically valid defect datasets with pixel-level annotations. The framework begins with the ab initio synthesis of defect layouts using controllable, physics-constrained mathematical morphology operations (erosion and dilation) applied to the original design-level layout. These synthesized layouts, together with their defect-free counterparts, are fabricated into physical samples via high-fidelity digital micromirror device (DMD)-based maskless lithography. Optical microscope images of the synthesized defect samples and their defect-free references are then compared to create consistent defect delineation annotations. Using this methodology, we constructed a comprehensive dataset of 3,530 Optical microscope images containing 13,365 annotated defect instances including four classes: bridge, burr, pinch, and contamination. Each defect instance is annotated with a pixel-accurate segmentation mask, preserving full contour and geometry. The segmentation-based Mask R-CNN achieves AP@0.5 of 0.980, 0.965, and 0.971, compared with 0.740, 0.719, and 0.717 for Faster R-CNN on bridge, burr, and pinch classes, representing a mean AP@0.5 improvement of approximately 34\%. For the contamination class, Mask R-CNN achieves an AP@0.5 roughly 42\% higher than Faster R-CNN. These consistent gains demonstrate that our proposed methodology to generate defect datasets with pixel-level annotations is feasible for robust AI-based Measurement/Inspection (MI) in semiconductor fabrication.
\end{abstract}

\section{Introduction}
The relentless scaling of Complementary Metal-Oxide-Semiconductor (CMOS) technology has continued to reduce critical dimensions (CDs) and increase the complexity of the layout \cite{RN53,RN52}. As CDs approach the nanometer regime, optical lithography, the foundational technique for pattern transfer, encounters fundamental physical constraints governed by the Rayleigh criterion \(CD = k_1 \lambda / \mathrm{NA}\) \cite{RN54}. The discrepancy between the illumination wavelength \(\lambda\) and the target feature size exacerbates Optical Proximity Effects (OPE), leading to pronounced pattern distortions and substantial degradation in pattern fidelity \cite{RN1,RN2}. These distortions often manifest as manufacturing defects, including shorts and opens, in layout regions known as lithographic defects \cite{RN55,RN56,RN57}.

Lithographic defects refer to localized regions where the printed patterns deviate from the intended design beyond acceptable process tolerances, which may vary from several nanometers to tens of nanometers depending on the technology node and specific layer \cite{RN21,RN25}. Although not all defects immediately result in device failure, they represent critical yield detractors. Unmitigated defects can propagate through downstream steps such as etching and deposition, eventually causing severe defects that significantly reduce manufacturing yield \cite{RN58,RN59}. Consequently, fast and accurate defect prediction and detection have become essential components of modern process control and yield management \cite{RN9}.

Artificial Intelligence (AI), particularly Deep Learning (DL), has recently emerged as a powerful tool across the broad electronics manufacturing landscape, ranging from defect analysis in manufacturing process to performance prediction due to its capacity to learn complex spatial patterns and extract subtle features from layout or glass image data \cite{RN11,RN12,RN17,RN64}. However, the performance of these data-driven approaches is fundamentally limited by the scarcity of high-quality, accurately annotated training data \cite{RN25,RN11,RN45,RN41}. Real-world defect instances are rare, diverse, and difficult to capture at scale, making reliable annotations both labor-intensive and highly dependent on expert judgment \cite{RN60,RN61}.

Furthermore, access to industrial lithography inspection data is strictly limited, and virtually no public datasets exist for research purposes due to confidentiality and process-dependent variability. This scarcity prevents the development of standardized training and benchmarking protocols. Since the scale and diversity of datasets strongly influence the performance and robustness of deep-learning-based defect detection models, constructing a reproducible and physically meaningful dataset becomes essential \cite{RN65}.

Historically, two strategies have been adopted to mitigate this data scarcity. The first relies on computational lithography simulations to generate synthetic defect patterns \cite{RN2}. By modeling the optical system (e.g., Hopkins formulation), photoresist behavior, and etch processes, simulated substrate patterns can be produced to identify potential defects \cite{RN23,RN26}. Yet, this approach suffers from inherent limitations. A pronounced distribution gap exists between simulated data and real manufacturing conditions, which include tool drift, material variability, and stochastic fluctuations\cite{RN22}. Models trained exclusively on simulated data often suffer from substantial performance degradation when applied to real production environments. Furthermore, high-fidelity simulations are computationally expensive, restricting their scalability for large and diverse dataset generation.

More recently, generative models such as Generative Adversarial Networks (GANs) and Denoising Diffusion Probabilistic Models (DDPMs) have been explored as an alternative for defect data augmentation \cite{RN17,RN7,RN6,RN29}. Although these models can expand small real datasets by generating visually realistic patterns, their data-driven nature limits their ability to cover the full range of physically plausible defects. Additionally, they do not resolve the fundamental labeling challenge; they still require an initial set of consistent, well-annotated real data, whose acquisition depends heavily on expert judgment and is difficult to standardize across annotators \cite{RN43,RN44}.

\begin{figure}[H]
    \centering
    \includegraphics[width=1\linewidth]{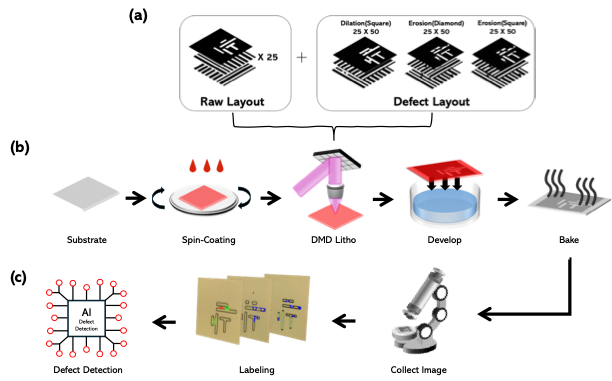}
    \caption{Schematic of the methodology for generating a physically grounded defect dataset. (a) Generation of design-level data, comprising 25 unique raw (defect-free) layouts and their corresponding defect-injected counterparts. Defects are systematically introduced via physics-constrained dilation and erosion operations. (b) High-fidelity physical replication of the designed layouts using DMD-based lithography. The process involves spin-coating AZ nLOF 2035 negative photoresist on a substrate, followed by DMD exposure, development, and a final baking step. (c) Ground truth generation and model training pipeline. Optical micrographs of the fabricated patterns are collected, and pixel-level ground truth is obtained by comparing the defect and raw layout images. This labeled dataset is then used to train an AI-based defect detection model.}
    \label{fig:figure1}
\end{figure}

To address the challenges, this study proposes a novel methodology for generating physically valid, pixel-level defect datasets by linking controllable design-level perturbations with high-fidelity physical replication. The proposed approach is based on the observation that a lithographic defect corresponds to a topological deviation from the intended design. Physically plausible defect patterns are systematically synthesized ab initio by applying controllable, physics-constrained erosion and dilation operations directly to the layout at the design stage. These benchmark layout pairs consisting of defect-free and perturbed variants are then faithfully reproduced using Digital Micromirror Device (DMD)-based maskless lithography. The Optical micrographs of the printed defect patterns are visually compared with their defect-free references, providing an objective and consistent reference for defect delineation and reducing annotator subjectivity and expert-to-expert variability. Figure~\ref{fig:figure1} provides a schematic overview of the complete end-to-end workflow, from digital layout manipulation to the final AI-ready dataset.

\section{Method}
This section provides a detailed exposition of the proposed framework for constructing a high-fidelity lithographic defect dataset. The core concept of the framework is to formulate a mathematical model for defect generation at the design-layout level and to reproduce these controlled perturbations physically through a systematic and reproducible workflow, ultimately enabling reliable annotation for AI-based inspection models. This section first introduces the theoretical formulation for defect synthesis, then describes the complete technical pipeline for dataset construction, and finally details the experimental setup, which is consistently used across dataset construction, model training, and performance evaluation.\\

\noindent\textit{\textbf{A. Theoretical Framework for Defect Synthesis}}\\

To generate diverse defect morphologies in a systematic and controllable manner, a morphology-based mathematical–physical model is established following the principles of Mathematical Morphology \cite{RN30}. In this formulation, a defect-free binary layout is represented as a set \(A \subset \mathbb{Z}^2\) on the two-dimensional integer lattice, where pixels with value 1 denote the foreground pattern (e.g., photoresist) and pixels with value 0 denote the background. Lithographic defects are modeled as localized morphological perturbations that alter the topology of the ideal layout \(A\).

These perturbations are introduced by applying a structuring element (SE) \(k_r\) of scale \(r\) at a target location \(t \in \mathbb{Z}^2\). A unified operator \(\sigma \in \{-1, +1\}\) characterizes the perturbation type, where \(\sigma = -1\) corresponds to erosion (representing pinching or open-circuit defects) and \(\sigma = +1\) corresponds to dilation (representing bridging or short-circuit defects). The perturbed layout \(A'\) is obtained through Minkowski operations:
\begin{equation}
  A' =
  \begin{cases}
    A \ominus k_r^{\,t}, & \sigma = -1 \quad \text{(erosion)} \\
    A \oplus k_r^{\,t},  & \sigma = +1 \quad \text{(dilation)} 
  \end{cases}
  \label{eq:minkowski}
\end{equation}
where \(k_r^{\,t}\) is the SE translated to location \(t\), and \(\oplus\) and \(\ominus\) denote Minkowski addition and subtraction, respectively.

To link this geometric perturbation to the final on-glass contour, a linear approximation model widely used in computational lithography \cite{RN1,RN2} is employed. The equivalent boundary displacement \(\Delta b_k(\mathbf{n})\) along the local boundary normal vector \(\mathbf{n}\) induced by the perturbation is expressed through the support function of the structuring element \(h_k(\mathbf{n})\):
\begin{equation}
  \Delta b_k(\mathbf{n}) = \sigma\, h_k(\mathbf{n})
  \label{eq:deltab}
\end{equation}
where the support function is given by
\begin{equation}
  h_k(\mathbf{n}) \equiv \sup_{\mathbf{x} \in k} \bigl\{ \mathbf{x} \cdot \mathbf{n} \bigr\}.
  \label{eq:support}
\end{equation}

This small boundary displacement on the Digital Micromirror Device (DMD) mask is amplified during optical projection by the Mask Error Enhancement Factor (MEEF), ultimately manifesting as an Edge Placement Error (EPE) on the substrate \cite{RN1,RN2}:
\begin{equation}
  \Delta \mathrm{EPE}(\mathbf{n}) = \mathrm{MEEF}(\mathbf{n}) \, \Delta b_k(\mathbf{n}).
  \label{eq:epe}
\end{equation}
Equations~\eqref{eq:minkowski}–\eqref{eq:epe} collectively establish the physical chain linking a digital-layout perturbation to boundary displacement and, ultimately, to the post-lithography contour error, thereby providing a principled and controllable basis for synthesizing physically meaningful defect morphologies.\\

\noindent\textit{\textbf{B. Defect Classification and Judgment Criteria}}\\

To objectively classify the synthesized lithographic defects, we define a set of criteria based on topological connectivity and morphological deformation. This formulation provides greater robustness than conventional approaches that rely solely on fixed geometric thresholds (e.g., minimum spacing).

The core idea is to quantify changes in the layout's connected components. In a 2D digital space, a connected component is defined as a set of foreground pixels such that any pair of pixels can be linked through an 8-connected path. Let \(k(A)\) denote the total number of connected components in a defect-free layout \(A\). After applying a morphological perturbation defined by the structuring element \(k_r\) and operator \(\sigma\), the perturbed layout becomes \(A'\). The resulting change in connectivity is quantified by
\begin{equation}
  \Delta k = k(A') - k(A).
  \label{eq:deltak}
\end{equation}

Based on the integer value of \(\Delta k\) and local contour characteristics, defects are categorized into three classes:
\begin{enumerate}
  \item \textbf{Pinch defect:} An erosion perturbation (\(\sigma = -1\)) may split a single connected component into multiple disconnected parts, resulting in \(\Delta k > 0\). To ensure physical realism and avoid idealized geometric cuts, the fractured contour is required to exhibit irregularity rather than a straight, artificially clean separation.
  \item \textbf{Bridge defect:} A dilation perturbation (\(\sigma = +1\)) may merge two or more previously isolated components into a single connected structure, leading to \(\Delta k < 0\). This topological signature corresponds directly to a short-circuit condition.
  \item \textbf{Burr defect:} A dilation perturbation may also produce a geometric protrusion that locally deforms the contour without fully connecting adjacent components. In this case, the connectivity remains unchanged (\(\Delta k = 0\)). Such protrusions represent pre-bridging states and are identified through significant irregular local deformation.
\end{enumerate}

\begin{figure}[H]
    \centering
    \includegraphics[width=\linewidth]{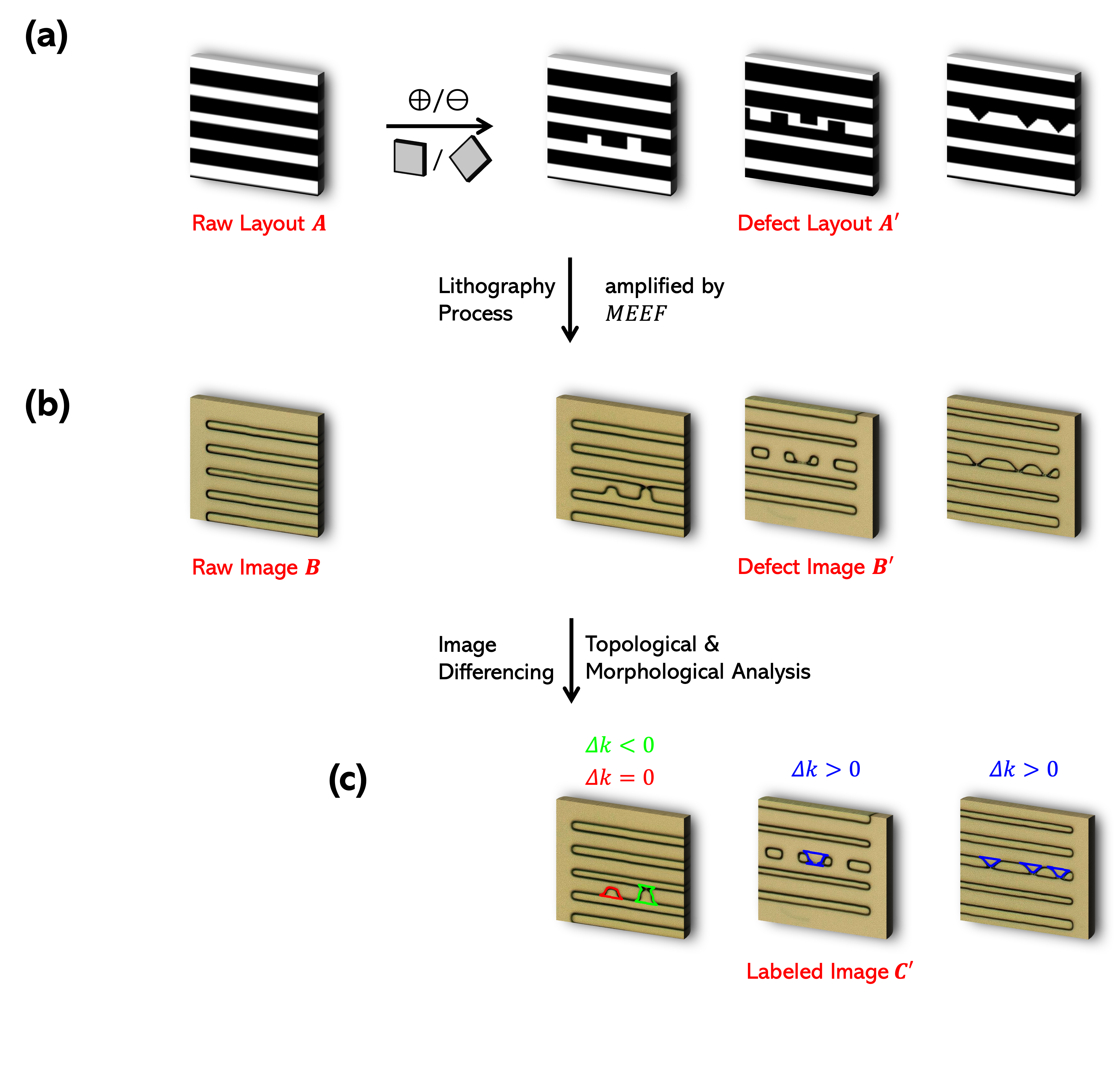}
    \caption{Conceptual framework for defect synthesis, physical manifestation, and classification.
    (a) Design-level synthesis. A defect-free raw layout \(A\) is transformed into various defect layouts \(A'\) through mathematical morphology operations (Minkowski erosion \(\ominus\) or dilation \(\oplus\)) using a designated structuring element (SE). 
    (b) Physical replication. The layouts \(A\) and \(A'\) are fabricated on the substrate using the lithography process, yielding a defect-free image \(B\) and corresponding defect images \(B'\). The designed perturbations are physically amplified by the Mask Error Enhancement Factor (MEEF), producing distinct defect images \(B'\) (bridge, burr, pinch) alongside the defect-free raw image \(B\). 
    (c) Ground-truth annotation. Image differencing between \(B\) and \(B'\), combined with topological and morphological analysis (evaluation of \(\Delta k = k(A') - k(A)\) and contour deformation), provides objective cues for consistent pixel-level annotation, resulting in labeled images \(C'\).}
    \label{fig:figure5}
\end{figure}

The overall conceptual framework linking digital defect synthesis, its physical manifestation on the printed substrate, and the subsequent classification logic is illustrated in Figure~\ref{fig:figure5}.

\noindent\textit{\textbf{C. Layout Library and Defect Injection Strategy}}\\

To ensure adequate diversity and practical relevance, we systematically designed a library of \(128 \times 128\)-pixel defect-free base layouts comprising two categories of patterns. The first category, \textit{complex composite patterns}, drew inspiration from academic benchmarks (e.g., ICCAD contests) \cite{RN62} and incorporated combinations of horizontal, vertical, and \(45^{\circ}\) diagonal features to reflect the geometric variability observed in real circuit environments. The second category, \textit{fundamental routing patterns}, consisted of densely packed horizontal-only or vertical-only line arrays, also defined on the same \(128 \times 128\)-pixel grid, enabling targeted analysis of orientation-dependent failure modes.

From the full layout library, 25 representative base patterns were selected for defect injection: 15 complex composite patterns, 5 horizontal-line patterns, and 5 vertical-line patterns. A constrained random sampling strategy was employed to inject defects onto each base layout, generating a balanced and diverse population of defects. For each of the 25 base layouts, a total of 150 defects were synthesized across three groups: (1) 50 pinch defects induced using axis-aligned square SEs; (2) 50 pinch defects induced using diamond-shaped SEs, producing gradual line-thinning profiles reflective of process-sensitive interconnects; (3) 50 bridging-type defects induced using axis-aligned square SEs.

\begin{figure}[H]
    \centering
    \includegraphics[width=\linewidth]{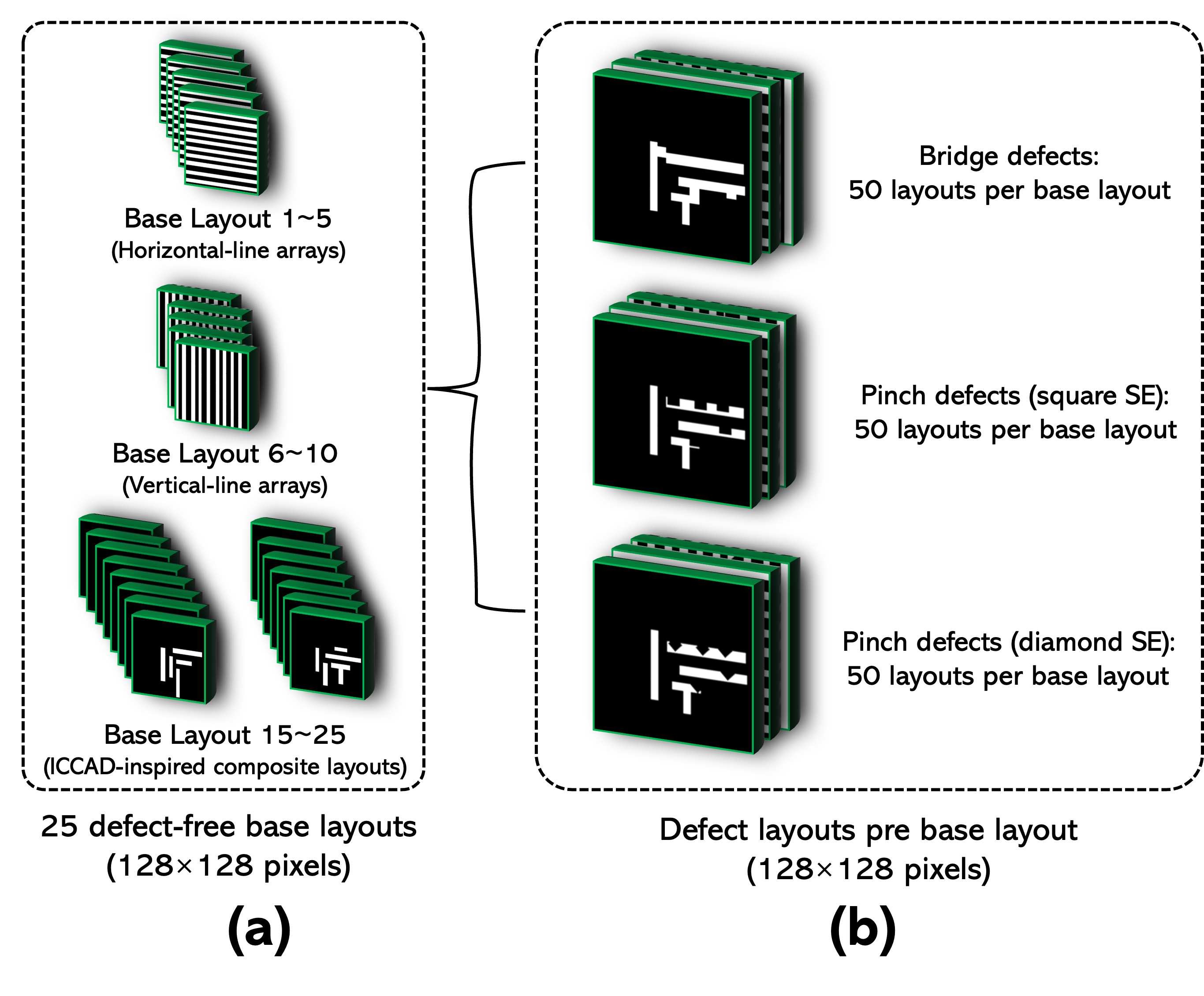}
    \caption{Layout library and defect injection strategy. 
    (a) Defect-free base layout library comprising 5 horizontal-line layouts, 5 vertical-line layouts, and 15 ICCAD-inspired composite layouts (25 base layouts in total, all defined on a \(128 \times 128\)-pixel grid). (b) Morphology-based defect injection scheme. For each base layout, 150 defect variants are synthesized using three defect groups: 50 bridge defects generated by dilation with a square structuring element (SE), 50 pinch defects generated by erosion with a square SE, and 50 pinch defects generated by erosion with a diamond-shaped SE.}
    \label{fig:figure6}
\end{figure}

\noindent\textit{\textbf{D. Experimental Workflow: From Layout to Labeled Image}}\\

The transformation of the digital layout library into a physically fabricated and consistently annotated image dataset followed a sequential experimental workflow. The process began with substrate preparation, in which 100-mm ITO-coated glass substrates were cleaned using ultrasonic agitation in acetone and isopropyl alcohol (IPA), rinsed with deionized (DI) water, and dried with high-purity nitrogen (\(\mathrm{N_2}\)).

A negative photoresist (Merck Performance Materials, AZ nLOF 2035) was then deposited via a two-stage spin-coating process (1500 rpm for 3 s followed by 3500 rpm for 40 s), yielding a uniform resist thickness of approximately \(4.0~\mu\mathrm{m}\). The coated substrates were soft-baked on a hotplate at \(110~^{\circ}\mathrm{C}\) for 60 s. Pattern exposure was performed using a DMD-based maskless lithography system (Paragon Ultra 80, Orbotech) equipped with a \(355~\mathrm{nm}\) light source \cite{RN13,RN14}. All layout pairs were exposed with a fixed dose of \(240~\mathrm{mJ/cm^2}\) and identical focus settings to ensure that any observed pattern variations arose solely from the designed perturbations.

Following exposure, development was conducted in a 2.38\% TMAH solution for 60 s, after which the substrates were hard-baked at \(110~^{\circ}\mathrm{C}\) for 60 s to enhance mechanical and chemical stability. High-resolution optical micrographs were subsequently captured using an optical microscope (KEYENCE, VHX-6000) with up to \(1500\times\) total magnification. Each image was centrally cropped to \(700 \times 700\) pixels to standardize its field of view.

Pixel-level ground-truth masks were obtained through a manual annotation process aided by image registration and side-by-side visual comparison between each raw image and its corresponding defect image. These aligned image pairs provided objective cues for accurately delineating defect regions while mitigating annotator subjectivity and inconsistency.\\

\noindent\textit{\textbf{E. Validation: Training and Evaluation}}\\

To assess the quality and generalizability of the constructed dataset, benchmarking was performed using representative models from two dominant computer-vision paradigms: instance-segmentation and object-detection \cite{RN27}. All training and evaluation experiments were conducted on a Dell Precision 7960 Tower workstation equipped with an Intel(R) Xeon(R) w9-3495X CPU, 1~TB RAM, and four NVIDIA RTX 6000 Ada Generation GPUs. The software environment consisted of CUDA~11.8 and PyTorch~2.1.1. The dataset was partitioned into approximately \(80\%\) training, \(10\%\) validation, and \(10\%\) testing, while ensuring mutual exclusivity across designs and exposure sessions to prevent data leakage. Preprocessing workflows and training budgets were standardized across models to enable a fair comparison.\\

\textit{\textbf{E-1. Training and Evaluation Protocol}}\\

All models were trained under a unified protocol. The default input resolution was \(704 \times 704\) pixels. Training was conducted for 100 epochs with a batch size of 4 and gradient accumulation of 4 (effective batch size 16). Optimization employed stochastic gradient descent (SGD) with an initial learning rate of 0.005, momentum of 0.937, and weight decay of \(5 \times 10^{-4}\). A linear warm-up phase of 3 epochs preceded MultiStepLR learning-rate decays at epochs 60 and 90 with \(\gamma = 0.1\).

To minimize confounding effects from aggressive augmentation, only horizontal flip (probability 0.5) and mild scaling (approximately \(\pm 10\%\)) were applied; mosaic, mixup, and copy-paste were disabled. All experiments utilized fp16 automatic mixed precision (AMP). Backbone weights were initialized from pretrained checkpoints (COCO-seg for YOLOv8-seg; ImageNet for ResNet-50 and VGG-16), and the number of data-loading workers was set to 2. Throughput (FPS) was measured on the same GPU with batch size 1, fp16 enabled, and no test-time augmentation (TTA), averaged over all 356 test images.\\

\textit{\textbf{E-2. Metrics and Matching Rules}}\\

The Intersection over Union (IoU) between a ground-truth box \(A\) and a predicted box \(B\) is defined as:\\\\
\begin{equation}
  \mathrm{IoU}(A, B) = \frac{|A \cap B|}{|A \cup B|}
  \label{eq:iou}
\end{equation}\\

Precision, the accuracy of the predictions, is calculated as the fraction of predicted instances that are actual defects:\\\\
\begin{equation}
  \mathrm{Precision} = \frac{\mathrm{TP}}{\mathrm{TP} + \mathrm{FP}}
  \label{eq:precision}
\end{equation}\\

Recall is the model's ability to find all actual defects and is calculated as:\\\\
\begin{equation}
  \mathrm{Recall} = \frac{\mathrm{TP}}{\mathrm{TP} + \mathrm{FN}}
  \label{eq:recall}
\end{equation}\\

The primary performance metric is mean Average Precision at IoU~0.5 (mAP@0.5). For each class, Average Precision (AP) is computed as the area under the precision–recall curve at \(\mathrm{IoU} = 0.5\), and mAP is obtained by averaging AP across all classes.\\

\textit{\textbf{E-2. Metrics and Matching Rules}}\\

Architectural characteristics that materially influence accuracy or speed under the unified training protocol are summarized as follows. YOLOv8-seg is a single-stage dense predictor employing a C2f backbone, an SPPF neck, and a prototype-based instance-segmentation head; cosine learning-rate scheduling is enabled in the Ultralytics implementation, while all other settings follow the unified protocol. The N/S/M/L variants scale depth and width to balance throughput and accuracy. Faster R-CNN (ResNet-50-FPN) is a two-stage detector that generates region proposals using a region proposal network (RPN), applies RoI Align, and performs classification and box regression in the detection head. To ensure consistency, both the minimum and maximum input sizes were fixed to 704 pixels. Mask R-CNN (ResNet-50-FPN, MMDetection) augments Faster R-CNN with a fully convolutional mask head for instance-segmentation. The preprocessing pipeline applied Resize(keep ratio) followed by Pad(704, 704). Evaluations were conducted using both bounding-box (bbox) and segmentation COCO metrics. Default MMDetection post-processing used a score threshold of 0.05, bounding-box NMS at \(\mathrm{IoU} = 0.5\), and mask binarization at 0.5.

\section{Result}

\noindent\textit{\textbf{A. Defect Dataset: Physical Generation and Outcomes}}\\

\textit{\textbf{A-1. From Layouts to Post-lithography Images}}\\

In accordance with the workflow in Figure~\ref{fig:figure1}, the defect dataset was generated through a physical pipeline that projects design-level morphological perturbations into printed resist contours via the lithographic system. Boundary displacements introduced in the layout by structuring-element (SE)-based erosion or dilation operations, denoted by \(\Delta b_k(\mathbf{n})\) in Eq.~(2), are amplified during optical projection by the Mask Error Enhancement Factor (MEEF) and manifest as Edge Placement Error (EPE) on the developed patterns according to Eq.~(4), under fixed dose and focus settings.

\begin{figure}[H]
    \centering
    \includegraphics[width=1\linewidth]{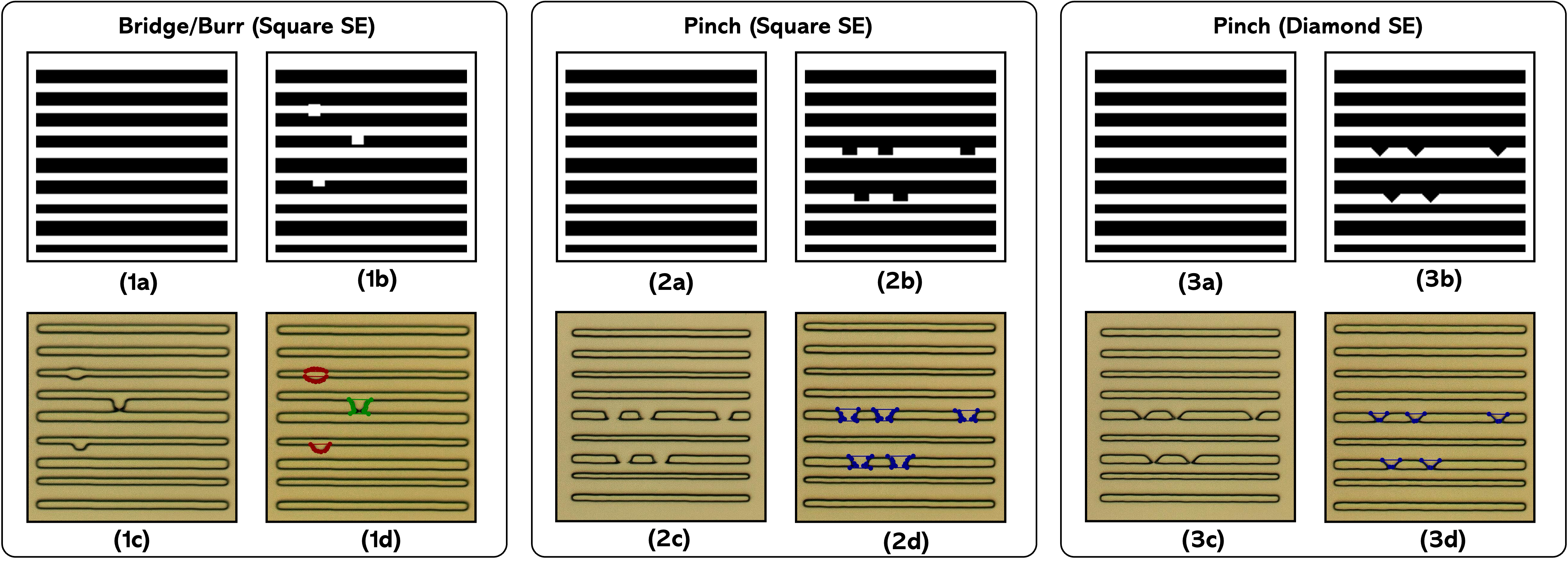}
    \caption{Examples across the design-to-image pipeline. (1a-1d) Bridge/Burr; (2a-2d) Pinch with a square Structuring Element (SE); (3a-3d) Pinch with a diamond SE. Within each row, panels correspond to: (a) raw layout (defect-free binary pattern); (b) SE-induced defect layout, where the boundary perturbation is defined by Eq. (2); (c) optical micrograph acquired under standardized illumination and magnification, showing deviations consistent with Mask Error Enhancement Factor (MEEF)-amplified perturbations; (d) pixel-level mask overlay with class-consistent color mapping. Equal magnification and fixed scale bars are used throughout}
    \label{fig:figure2}
\end{figure}

Empirical measurements indicate that the magnitude of the computed design-level displacement \(\lvert \Delta b_k \rvert\) is positively correlated with the measured \(\lvert \Delta \mathrm{EPE} \rvert\) on standardized microscope images. Variations in SE geometry and SE scale further produce systematic shifts in mask-area deformation and necking-width distributions. Representative examples spanning the entire design-to-image pipeline are provided in Figure~\ref{fig:figure2}, illustrating the transfer of SE-induced perturbations into their printed counterparts. These observations confirm that the intended design-time perturbations are consistently transferred to the printed contours, supporting both the physical plausibility of the annotated defect instances and the generalization performance observed in downstream benchmarks.

\textit{\textbf{A-2. Dataset Statistics of Post-lithography Image}}\\

Table~\ref{tab:table1} summarizes the instance-level statistics of the curated defect dataset across the train/validation/test splits. In total, the dataset consists of 3{,}530 images and 13{,}365 annotated instances spanning four classes: Bridge, Burr, Pinch, and Contamination. The Contamination class consists of naturally occurring defects rather than synthetically introduced ones, leading to a class imbalance with substantially fewer Contamination instances. The dataset was partitioned into approximately \(80\%\) training, \(10\%\) validation, and \(10\%\) testing at the image level. To prevent leakage, the splits were made mutually exclusive by both designs and exposure sessions. All counts reflect post-processing and quality-controlled annotations obtained under standardized imaging conditions.

\begin{table}[H]
    \centering
    \captionsetup{justification=centering}
    \setlength{\tabcolsep}{7pt} 
    \renewcommand{\arraystretch}{2} 
    \caption{Instance counts per split and class (Bridge, Burr, Pinch, and Contamination).\\}
    \label{tab:table1}
\begin{tabular}{ccccccc}
    \hline
    &        & \multicolumn{5}{c}{Class} \\
    \cline{3-7}
    Dataset & Total Images & Bridge & Burr & Pinch & Contamination & Total \\
    \hline
    Train & 2{,}823 & 1{,}193 & 1{,}674 & 7{,}769 & 56 & 10{,}692 \\
    Val   & 351     & 145     & 217     & 954     & 3  & 1{,}319 \\
    Test  & 356     & 155     & 206     & 981     & 12 & 1{,}354 \\
    \hline
    Total & 3{,}530 & 1{,}493 & 2{,}097 & 9{,}704 & 71 & 13{,}365 \\
    \hline
\end{tabular}
\end{table}

\begin{figure}[H]
    \centering
    \includegraphics[width=1\linewidth]{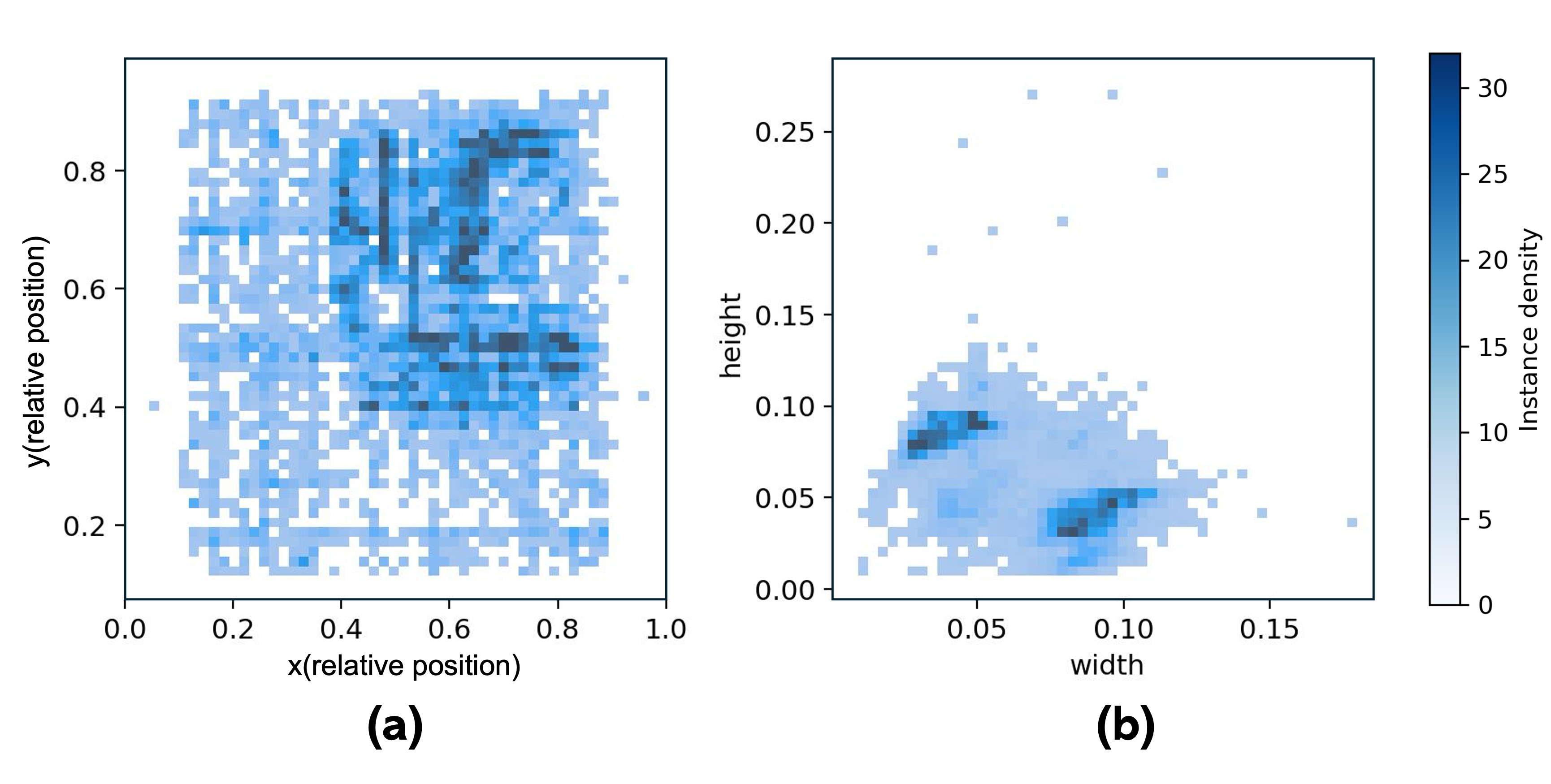}
    \caption{Dataset statistics. (a) Spatial density map of annotated instances showing a slight central positional bias. (b) Distribution of instance mask size (percentage of total image size). Statistics computed on the training dataset under the preprocessing conditions described in Methods.}
    \label{fig:figure3}
\end{figure}

The spatial and morphological distribution of annotated defects was analyzed to characterize the dataset. Figure~\ref{fig:figure3}(a) presents the spatial density map of annotated instances, showing a mild central positional bias that aligns with the optical-field uniformity of the lithographic system. Figure~\ref{fig:figure3}(b) shows the distribution of instance mask sizes, demonstrating that defect sizes span a broad range of scales. Together, these statistics confirm that the dataset exhibits substantial geometric diversity, providing comprehensive coverage of fine-scale and mid-scale defect manifestations.\\

\noindent\textit{\textbf{B. Defect Detection Validation}}\\

\textit{\textbf{B-1. Quantitative Results}}\\

Each defect instance in the dataset is annotated by a pixel-accurate mask that traces the printed defect contour. This representation is suited for instance-segmentation models, which aim to recover the full spatial extent and topology of each defect. To enable a fair comparison with object-detection architectures, axis-aligned bounding boxes were algorithmically derived from the segmentation masks by taking the tightest enclosing rectangle around each instance. Consequently, both segmentation and detection models are trained from the same underlying physical ground truth but differ in the granularity of their predicted outputs: segmentation models estimate dense masks, whereas detection models only regress bounding boxes.

On this unified annotation, two categories of models were evaluated:
\begin{enumerate}
    \item \textbf{Instance-segmentation models} (Mask R-CNN, YOLOv8-seg), which directly predict per-pixel masks for each defect instance; and
    \item \textbf{Object-detection models} (Faster R-CNN, YOLOv8-det), which predict bounding boxes and class labels only.
\end{enumerate}

For a consistent quantitative comparison, all methods were trained under identical settings, including dataset splits, data preprocessing, optimization hyperparameters, and training budget, and were evaluated using detection-style metrics (per-class AP@0.5) computed on bounding boxes. For segmentation models, the predicted masks were converted to tight boxes before evaluation. The per-class AP@0.5 results in Table~\ref{tab:table2} show that instance-segmentation models consistently provide an advantage over object-detection models on proposed dataset.

The segmentation-based Mask R-CNN attains per-class AP@0.5 values of 0.980, 0.965, 0.971, and 0.336 for bridge, burr, pinch, and contamination classes, respectively, compared with 0.740, 0.719, 0.717, and 0.237 for the detection-based Faster R-CNN. The mean AP@0.5 over all four defect classes (mAP@0.5) is 0.813 for Mask R-CNN and 0.603 for Faster R-CNN, corresponding to an approximate 35\% relative gain across the four classes. YOLOv8-seg and YOLOv8-det achieve nearly identical AP@0.5 on the bridge, burr, and pinch classes (differences below 1\%), whereas YOLOv8-seg attains about 48\% higher AP@0.5 on the contamination class, achieving 0.228 compared with 0.154 for YOLOv8-det.

These trends indicate that the benefit of instance-segmentation models is class dependent. For contamination, which occurs naturally in lithography and subsequent processes, instance-segmentation models still maintain a noticeable advantage over their detection counterparts despite overlapping with surrounding layout patterns. Instance-segmentation models, trained on pixel-accurate masks, can capture these geometry-dependent properties of rare and small defect more effectively than object-detection models trained with box-only labels.

\begin{table}[H]
    \centering
    \captionsetup{justification=centering}
    \setlength{\tabcolsep}{8pt} 
    \renewcommand{\arraystretch}{2} 
    \caption{Performance of instance-segmentation and object-detection models on the test set.\\}
    \label{tab:table2}
    \begin{tabular}{c c c c c c c}
        \hline
        & & & \multicolumn{4}{c}{AP@0.5} \\
        \cline{4-7}
        Type & Model & Backbone & Bridge & Burr & Pinch & Contamination \\
        \hline
        \multirow{2}{*}{\shortstack{Instance-\\Segmentation\\Model}} 
            & YOLOv8-seg & \shortstack{C2f-SPP}    & 0.995 & 0.989 & 0.991 & 0.228 \\
            & Mask R-CNN & \shortstack{ResNet-50} & 0.980 & 0.965 & 0.971 & 0.336 \\
        \hline
        \multirow{2}{*}{\shortstack{Object-\\Detection\\Model}} 
            & YOLOv8-det   & \shortstack{C2f-SPP}    & 0.989 & 0.983 & 0.990 & 0.154 \\
            & Faster R-CNN & \shortstack{ResNet-50}  & 0.740 & 0.719 & 0.717 & 0.237 \\
        \hline
    \end{tabular}
\end{table}

\textit{\textbf{B-2. Qualitative Results}}\\

\begin{figure}[H]
    \centering
    \includegraphics[width=1\linewidth]{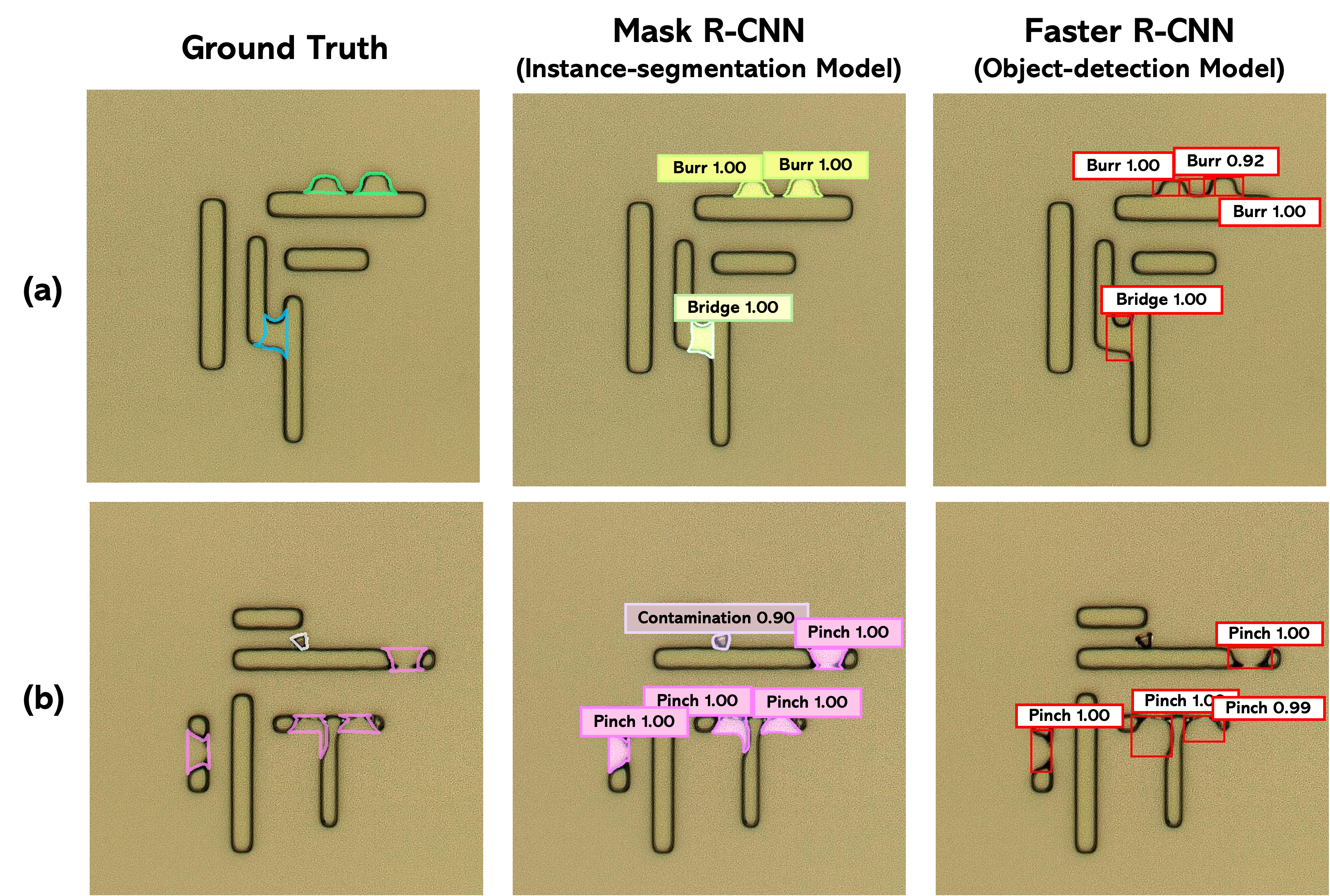}
    \caption{Qualitative detection results from instance-segmentation and object-detection models. (a) and (b) correspond to different defect layouts. Ground-truth defect masks are overlaid on the Optical micrographs; blue, green, pink, and white denote bridge, burr, pinch, and contamination classes, respectively. For predicted instances, the numbers next to each class label indicate the detector’s confidence score (maximum softmax probability for the predicted class).}
    \label{fig:figure4}
\end{figure}

Figure~\ref{fig:figure4} compares representative qualitative predictions from the instance-segmentation models and the object-detection models on the proposed defect dataset. The segmentation masks closely follow the ground-truth annotations across diverse geometries and resist contrast conditions, indicating accurate boundary localization and robust preservation of defect morphology. Segmentation masks detect pixel-level shapes, whereas detection boxes provide compact box-level localization of the same defects, showing that, in typical cases, both models identify nearly the same defect regions while offering different levels of spatial detail.

In Figure~\ref{fig:figure4}(a), Faster R-CNN, the object-detection model, produces an additional detection in a region that is not annotated in the ground truth, indicating a false positive on the background. In Figure~\ref{fig:figure4}(b), the same model fails to detect a contamination instance that is annotated in the ground truth. These qualitative failure modes are consistent with the quantitative per-class performance: the segmentation-based Mask R-CNN attains AP@0.5 values of 0.980, 0.965, 0.971, and 0.336 for bridge, burr, pinch, and contamination classes, respectively, compared with 0.740, 0.719, 0.717, and 0.237 for the detection-based Faster R-CNN. These correspond to approximate relative gains of 32\%, 34\%, 35\%, and 42\% on bridge, burr, pinch, and contamination classes. The mean AP@0.5 over all four defect classes (mAP@0.5) is 0.813 for Mask R-CNN and 0.603 for Faster R-CNN, representing an overall relative improvement of about 35\%. These results indicate that instance-segmentation models not only sharpen boundary localization on topology-driven defects but also substantially reduce both spurious background activations and missed detections for small, irregular contamination defects.

These observations suggest that pixel-level prediction in instance-segmentation models is more robust than bounding-box-based detection for small, irregular contamination defects and that instance-segmentation architectures provide not merely a more detailed label format but a more appropriate prediction mechanism overall for the proposed dataset. This is particularly important because, in defect inspection in the lithography process, the critical information lies in the exact contour shape and local geometry of each defect rather than only in its coarse location.

\section{Conclusion}

This work presents a physically grounded framework for constructing large-scale lithographic defect datasets that integrate controllable design-level perturbations with high-fidelity physical replication. By formulating defects through Mathematical Morphology and transferring these perturbations to printed contours via DMD-based maskless lithography, the proposed methodology establishes a direct and verifiable linkage between digital defect synthesis and on-substrate manifestations. This approach addresses the long-standing bottleneck of scarce, inconsistently annotated real-world training data, replacing purely simulation-based approximation data with the proposed defect shape dataset that are realized through actual exposure, development, and imaging.

The resulting dataset comprises 3,530 Optical micrographs and 13,365 annotated defect instances across four defect classes, demonstrating substantial geometric diversity and strong alignment between design-time perturbations and printed outcomes. A key characteristic of the dataset is that every defect is annotated with a pixel-accurate segmentation mask rather than a bounding box, allowing learning algorithms to exploit detailed contour shape, local topology, and subtle line-thinning or protrusion features that are critical for lithographic defect behavior. 

Under a unified evaluation protocol based on bounding-box metrics, the segmentation-based Mask R-CNN achieves per-class AP@0.5 values of 0.980, 0.965, 0.971, and 0.336 on the bridge, burr, pinch, and contamination classes, respectively, compared with 0.740, 0.719, 0.717, and 0.237 for the detection-based Faster R-CNN. These correspond to relative AP@0.5 gains of approximately 32\%, 34\%, and 35\% on bridge, burr, and pinch classes and about 42\% on contamination class. As a result, the mean AP@0.5 (mAP@0.5) is 0.813 for Mask R-CNN and 0.603 for Faster R-CNN, corresponding to an overall improvement of about 35\%. These gains indicate that mask-level supervision translates into more complete and precise defect coverage even when performance is measured at the box level. These results confirm that the physically grounded dataset effectively supports the training of robust, high-performance AI models, exhibiting strong generalization across architectures and defect morphologies.

From a practical standpoint, the DMD-based maskless lithography platform is pivotal to the scalability and flexibility of the proposed framework. Because exposure patterns can be reprogrammed electronically, a wide variety of synthesized layouts including dense line arrays, composite routing structures, and ICCAD-inspired patterns can be realized without the need for new photomasks. This enables systematic exploration of diverse pattern contexts and process windows while preserving the physical realism of the resulting defect shapes, which naturally inherit optical proximity effects, resist blur, and process variability from the underlying lithography system. The framework introduced in this study provides a scalable and systematic pathway for generating physically representative training data for AI-driven lithography inspection. By enabling controllable, reproducible, and quantitatively verifiable defect generation, it establishes a foundation for developing more reliable inspection models capable of adapting to real process variability. Future extensions may incorporate a broader spectrum of defect types, additional process layers, and more complex layouts, as well as assessing the transferability of trained models to inline or near-inline inspection environments. This dataset-generation paradigm is poised to support next-generation process control strategies and accelerate the deployment of AI-based inspection systems in semiconductor manufacturing. The detailed mathematical formulation of defect synthesis, layout library design, lithography conditions, and benchmarking protocol is provided in the Methods section.

\section{Acknowledgments}

This work was supported by the Technology Innovation Program funded by the Ministry of Trade, Industry and Energy (MOTIE, Korea) [Project number:00438129].
\section{Data Availability}

The dataset generated in this study are available from the corresponding author upon reasonable request.
\section{Author Contributions}

Y. H. Hu designed the overall defect-generation framework, constructed the morphology-based lithographic dataset, performed the DMD-based fabrication and microscopic measurements, and wrote the main manuscript. J. Y. Kong jointly contributed to dataset construction, lithography experiments, data processing, and manuscript preparation. D. Y. Shin provided technical discussions during the course this project. J. K. Kim and K.-T. Kang supervised the project and provided conceptual guidance. All authors reviewed and approved the manuscript.
\section{Competing interests}

The authors declare no competing interests.

\bibliographystyle{unsrt}
\bibliography{ref}

\end{document}